\documentclass[11pt,a4paper]{article}
\usepackage[hyperref]{word_embed}
\usepackage{times}
\usepackage{latexsym}
\usepackage{amssymb}
\usepackage{comment}
\usepackage{commath}
\usepackage{amsmath}
\usepackage{url}
\usepackage{bm}
\usepackage{textcomp}

\usepackage[normalem]{ulem}
\useunder{\uline}{\ul}{}

\DeclareMathOperator{\bx}{\mathbf{x}}
\DeclareMathOperator{\bmu}{\bm{\mu}}

\usepackage{url}

\aclfinalcopy 

\title{Learning Multi-Sense Word Distributions using Approximate Kullback-Leibler Divergence}

\author{P.~Jayashree, 
  Ballijepalli Shreya, 
  and P.K.~Srijith \\
  Department of Computer Science and Engineering, Indian Institute of Technology,\\
  Hyderabad, India  \\
  {\tt \{cs16resch11002,cs15btech11009,srijith\}@iith.ac.in} \\}

\date{}

\begin{document}
\maketitle
\begin{abstract}
  Learning word representations has garnered greater attention in the recent past due to its diverse text applications. Word embeddings encapsulate the syntactic and semantic regularities of sentences. Modelling word embedding as multi-sense gaussian mixture distributions, will additionally capture uncertainty and polysemy of words.  We propose to learn the Gaussian mixture representation of words using a Kullback-Leibler (KL) divergence based objective function. The KL divergence based energy function provides a better distance metric which can effectively capture entailment and distribution similarity among the words. Due to the intractability of KL divergence for Gaussian mixture, we go for a KL approximation between Gaussian mixtures. We perform qualitative and quantitative experiments on benchmark word similarity and entailment datasets which demonstrate the effectiveness of the proposed approach.\\
\end{abstract}

\section{Introduction}

Language modelling in its inception had one-hot vector encoding of words. However, it captures only alphabetic ordering but not the word semantic similarity. 
Vector space models helps to learn word representations in a lower dimensional space and also captures semantic similarity. Learning word embedding aids in natural language processing tasks such as question answering and reasoning \cite{choi2018quac}, stance detection \cite{augenstein2016stance}, claim verification \cite{hanselowski2018ukp}.

Recent models \citep{mikolov2013efficient,bengio2003neural} work on the basis that words with similar context share semantic similarity. \citet{bengio2003neural} proposes a neural probabilistic model which models the target word probability conditioned on the previous words using a recurrent neural network. \emph{Word2Vec} models \citep{mikolov2013efficient} such as continuous bag-of-words (CBOW) predict the target word given the context, and skip-gram model works in reverse of predicting the context given the target word. While,  \emph{GloVe} embeddings were based on a Global matrix factorization on local contexts \citep{pennington2014glove}. However, the aforementioned models do not handle words with multiple meanings (polysemies).

\citet{huang2012improving} proposes a neural network approach considering both local and global contexts in learning word embeddings (point estimates). Their multiple prototype model handles polysemous words by providing apriori heuristics about word senses in the dataset. \citet{tian2014probabilistic} proposes an alternative to handle polysemous words by a modified skip-gram model and EM algorithm. \citet{neelakantan2015efficient} presents a non-parametric based alternative to handle polysemies. 
However, these approaches fail to consider entailment relations among the words.

\citet{vilnis2014word} 
learn a Gaussian distribution per word using the expected likelihood kernel. 
However, for polysemous words, this may lead to word distributions with larger variances as it may have to cover various senses. 

\citet{athiwaratkun2017multimodal} proposes multimodal word distribution approach. It captures polysemy. However, the energy based objective function fails to consider asymmetry and hence entailment. Textual entailment recognition is necessary to capture lexical inference relations such as causality (for example, mosquito $\rightarrow$ malaria), hypernymy (for example, dog $\models$ animal) etc.




In this paper, we propose to obtain multi-sense word embedding distributions by using a variant of max margin objective based on the asymmetric KL divergence energy function to capture textual entailment. Multi-sense distributions are advantageous in capturing polysemous nature of words and in reducing the uncertainty per word by distributing it across senses. However, computing KL divergence between mixtures of Gaussians is intractable, and we use a KL divergence approximation based on stricter upper and lower bounds. While capturing textual entailment (asymmetry), we have also not compromised on capturing symmetrical similarity between words (for example, funny and hilarious) which will be elucidated in Section $3.1$. We also show the effectiveness of the proposed approach on the benchmark word similarity and entailment datasets in the experimental section.

\section{Methodology}
\subsection{Word Representation}
Probabilistic representation of words helps one model uncertainty in word representation, and polysemy.
Given a corpus $V$, containing a list of words each represented as $w$, the probability density for a word $w$ can be represented as a mixture of Gaussians with $C$ components~\citep{athiwaratkun2017multimodal}.
\begin{equation}
\label{eqn:mix_gauss_form}
 \begin{aligned}
    f_w(\bx) &= \sum_{j=1}^{C} p_{w,j} N(\bmu_{w,j}, \Sigma_{w,j}) ; \quad
    \sum_{j=1}^{C} p_{w,j} &= 1
 \end{aligned}
\end{equation}
Here, $p_{w,j}$ represents the probability of word $w$ belonging to the component $j$, $\bmu_{w,j}$ represents $D$ dimensional word representation corresponding to the $j^{th}$ component sense of the word $w$, and $\Sigma_{w,j}$ represents the uncertainty in representation for word $w$ belonging to component $j$. 

\section{Objective function}
The model parameters (means, covariances and mixture weights) $\theta$ can be learnt using a variant of max-margin objective \citep{joachims2002optimizing}.
\begin{equation}
\label{eqn:loss_fn}
 \begin{aligned}
    L_\theta(w,c,c') &= max(0, m - \log E_\theta (w,c) \\ &\qquad \qquad + \log E_\theta (w,c'))
     \end{aligned}
\end{equation}
Here $E_\theta (\cdot, \cdot)$ represents an  energy function which assigns a score to the pair of words, $w$ is a particular word under consideration, $c$ its positive context (same context), and $c'$ the negative context.   The objective aims to push the margin of  the difference between the energy function of a word $w$ to its positive context $c$ higher than its negative  context $c$  by a threshold of $m$. Thus, word pairs in the same context gets a higher energy than the word pairs in the dissimilar context. \citet{athiwaratkun2017multimodal} consider the energy function to be an expected likelihood kernel which is defined as follows.
\begin{equation}
\label{eqn:energy_EL}
     EL(w,c) = \int f_w(\bx) f_c(\bx) d\bx 
\end{equation}
This is similar to the cosine similarity metric over vectors and the energy between two words is maximum when they have similar distributions.  
But, the expected likelihood kernel is a symmetric metric which will not be suitable for capturing ordering among words and hence entailment. 

\subsection{Proposed Energy function}


As each word is represented by a mixture of Gaussian distributions, KL divergence is a better choice of energy function to capture distance between distributions. Since, KL divergence is minimum when the distributions are similar and maximum when they are dissimilar, energy function is taken as exponentiated negative KL divergence.
\begin{equation}
\label{eqn:energy1}
     E_\theta (w,c) = \exp( - KL (f_w(\bx) ||f_c(\bx)) ) 
\end{equation}

However, computing KL divergence between Gaussian mixtures is intractable and obtaining exact KL value is not possible.  One way of approximating the KL is by \emph{Monte-Carlo approximation} but it requires large number of samples to get a good approximation and is computationally expensive on high dimensional embedding space.   

Alternatively, \citet{hershey2007approximating} presents  a KL approximation between Gaussian mixtures where they obtain an upper bound through  product of Gaussian approximation method and a lower bound through variational approximation method. In \citep{durrieu2012lower}, the authors combine the lower and upper bounds from  approximation methods of \citet{hershey2007approximating} to provide a stricter bound on KL between Gaussian mixtures. Lets consider  Gaussian mixtures for the words $w$ and $v$ as follows. 
\begin{equation}
    \label{eqn:gmm}
    \begin{aligned}
    f_{w}(\bx) &= \sum_{i=1}^C p_{w,i}f_{w,i}(\bx) = \sum_{i=1}^C p_{w,i} N(\bx;\bmu_{w,i},\Sigma_{w,i}) \\
    f_{v}(\bx) &= \sum_{j=1}^C p_{v,j} f_{v,j}(\bx) = \sum_{j=1}^C p_{v,j} N(\bx;\bmu_{v,j},\Sigma_{v,j})
    \end{aligned} \nonumber
\end{equation}
The approximate KL divergence between the Gaussian mixture representations over the words $w$ and $v$ is shown in equation \ref{eqn:kl_approx}. More details on approximation is included in the Supplementary Material.
\begin{equation}
\label{eqn:kl_approx}
\begin{aligned}
& KL(f_{w}(\bx)||f_{v}(\bx)) \triangleq 
\\ & ({\sum_i} {p_{w,i}} \log{\frac{\sum_k {p_{w,k} EL_{ik}(w,w)}}{\sum_j p_{v,j} \exp({-KL(f_{w,i}(\bx)||f_{v,j}(\bx))})}} +\\ & {\sum_i} {p_{w,i}} \log{\frac{\sum_k
p_{w,k} \exp({-KL(f_{w,i} (\bx)||f_{w,k}(\bx))})}{\sum_j {p_{v,j}} EL_{ij}(w,v)}})/2
\end{aligned}
\end{equation}
where $EL_{ik}(w,w) = \int f_{w,i} (\bx) f_{w,k} (\bx) d\bx$ and $EL_{ij}(w,v) = \int f_{w,i} (\bx) f_{v,k} (\bx) d\bx$. Note that the expected likelihood kernel appears component wise inside the approximate  KL divergence derivation. 

One advantage of using KL as energy function is that it enables to capture asymmetry in entailment datasets. For eg., let us consider the words 'chair' with two senses as 'bench' and 'sling', and 'wood' with two senses as 'trees' and 'furniture'. The word chair ($w$) is entailed within wood ($v$), i.e.  chair $\models$ wood. Now, minimizing the KL divergence necessitates maximizing $\log{\sum_j p_{v,j} \exp({-KL(f_{w,i} (\bx)||f_{v,j}(\bx))})}$ which in turn \textbf{minimizes $KL(f_{w,i}(\bx)||f_{v,j}(\bx))$}. This will result in the support of the $i^{th}$ component of $w$ to be within the $j^{th}$ component of $v$, and holds for all component pairs leading to the \textbf{entailment of $w$ within $v$}. Consequently, we can see that bench $\models$ trees, bench $\models$ furniture, sling $\models$ trees, and sling $\models$ furniture. Thus, it introduces lexical relationship between the senses of child word and that of the parent word. Minimizing the KL also necessitates maximizing $\log{\sum_j {p_{v,j}} EL_{ij}(w,v)}$ term for all component pairs among $w$ and $v$. This is similar to \textbf{maximizing expected likelihood kernel, which brings  the means of $f_{w,i}(\bx)$ and $f_{v,j}(\bx)$ closer} (weighted by their co-variances) as discussed in \citep{athiwaratkun2017multimodal}. Hence, the proposed approach captures the \textbf{best of both worlds}, thereby catering to both word similarity and entailment.

We also note that minimizing the KL divergence necessitates minimizing $\log{\sum_k
p_{w,k} \exp({-KL(f_{w,i}||f_{w,k})})}$ which in turn maximizes $KL(f_{w,i}||f_{w,k})$. This prevents the different mixture components of  a word converging to single Gaussian and encourages capturing different possible senses of the word.  The same is also achieved by minimizing $\sum_k {p_{w,k}} EL_{ik}(w,w)$ term and act as a regularization term which promotes diversity in learning senses of a word.

\section{Experimentation and Results}
We train our proposed model \textbf{GM$\_$KL} (Gaussian Mixture using KL Divergence) on the Text8 dataset \citep{mikolov2014learning} which is a pre-processed data of $17M$ words from wikipedia. Of which, $71290$ unique and frequent words are chosen using the subsampling trick in \cite{mikolov2013distributed}. We compare \textbf{GM$\_$KL} with the previous approaches \textbf{w2g} \citep{vilnis2014word} ( single Gaussian model) and \textbf{w2gm} \citep{athiwaratkun2017multimodal} (mixture of Gaussian model with expected likelihood kernel). 
For all the models used for experimentation, the embedding size ($D$) was set to $50$, number of mixtures to $2$, context window length to $10$, batch size to $128$. The word embeddings were initialized using a uniform distribution in the range of $[-\sqrt{\frac{3}{D}}$, $\sqrt{\frac{3}{D}}]$ such that the expectation of variance is $1$ and mean $0$ \citep{cun1998efficient}. One could also consider initializing the word embeddings using other contextual representations such as BERT \citep{devlin2018bert} and ELMo \citep{peters2018deep} in the proposed approach. In order to purely analyze the performance of $\emph{GM\_KL}$ over the other models, we have chosen initialization using uniform distribution for experiments. For computational benefits, diagonal covariance is used similar to \citep{athiwaratkun2017multimodal}. Each mixture probability is constrained in the range $[0,1]$, summing to $1$ by optimizing over unconstrained scores in the range $(-\infty,\infty)$ and converting scores to probability using softmax function. The mixture scores are initialized to $0$ to ensure fairness among all the components. The threshold for negative sampling was set to $10^{-5}$, as recommended in \cite{mikolov2013efficient}. Mini-batch gradient descent with Adagrad optimizer \citep{duchi2011adaptive} was used with initial learning rate set to $0.05$.

Table \ref{table:quali_res} shows the qualitative results of GM$\_$KL. Given a query word and component id, the set of nearest neighbours along with their respective component ids are listed.  For eg., the word `plane' in its $0^{th}$ component captures the `geometry' sense and so are its neighbours, and its $1^{st}$ component captures `vehicle' sense and so are its corresponding neighbours. Other words such as `rock' captures both the `metal' and `music' senses, `star' captures `celebrity' and `astronomical' senses, and `phone' captures `telephony' and `internet' senses.


\begin{table}[ht]
\caption{\textbf{Qualitative results of GM$\_$KL}}
\begin{tabular}{|l|l|p{53mm}|}
\hline
\textbf{Word} & \textbf{Co.} & \textbf{Nearest Neighbours}                                                                                                                                    \\ \hline
rock          & 0            & \begin{tabular}[c]{@{}l@{}}rock:0, sedimentary:0, molten:1,\\granite:0, felsic:0, carvings:1, \\kiln:1\end{tabular}            \\ \hline
rock          & 1            & \begin{tabular}[c]{@{}l@{}}rock:1, albums:0, rap:0, album:0,\\ bambaataa:0, jazzy:0, remix:0\end{tabular}                     \\ \hline
star          & 0            & \begin{tabular}[c]{@{}l@{}}star:0, hulk:0, sequel:0, godzilla:0,\\ishiro:0, finale:1, cameo:1\end{tabular}                      \\ \hline
star          & 1            & \begin{tabular}[c]{@{}l@{}}star:1, galactic:0, stars:1, galaxy:1,\\galaxy:0, sun:1, brightest:1, \end{tabular}          \\ \hline
phone         & 0            & \begin{tabular}[c]{@{}l@{}}phone:0, dialing:0, voip:1, \\channels:0,cable:1, telephone:1,\\caller:0\end{tabular}            \\ \hline
phone         & 1            & \begin{tabular}[c]{@{}l@{}}phone:1, gsm:1, ethernet:1, \\wireless:1, telephony:0,\\transceiver:0, gprs:0\end{tabular}                  \\ \hline
plane         & 0            & \begin{tabular}[c]{@{}l@{}}plane:0, ellipse:0, hyperbola:0, \\ tangent:0,axis:0, torus:0, convex:0, \end{tabular}              \\ \hline
plane         & 1            & \begin{tabular}[c]{@{}l@{}}plane:1, hijacked:1, sidewinder:0,\\takeoff:1, crashed:0, cockpit:1, \\pilot:1\end{tabular}       \\ \hline
\end{tabular}
\label{table:quali_res}
\end{table}

We quantitatively compare the performance of the  GM$\_$KL, w2g, and w2gm approaches on the SCWS dataset \citep{huang2012improving}. The dataset consists of $2003$ word pairs of polysemous and homonymous words with labels obtained by an average of $10$ human scores. The Spearman correlation between the human scores and the model scores are computed. To obtain the model score, the following metrics are used:
\begin{enumerate}
    \item \textbf{MaxCos}: Maximum cosine similarity among all component pairs of words $w$ and $v$:
    \begin{equation}
    \emph{MaxCos(w,v)} = \max_{i,j=1,2,...,C}\frac{\langle \mu_{w,i}, \mu_{v,j} \rangle}{\norm \mu_{w,i} \cdot \norm \mu_{v,j}} \nonumber \end{equation}
    \item \textbf{AvgCos}: Average component-wise cosine similarity between the words $w$ and $v$.
    \begin{equation}
    \emph{AvgCos(w,v)} = \frac{1}{C} \sum_{i=1}^{C} \sum_{j=1}^{C} \frac{\langle \mu_{w,i}, \mu_{v,j} \rangle}{\norm \mu_{w,i} \cdot \norm \mu_{v,j}} \nonumber \end{equation}
    \item \textbf{KL$\_$approx}: Formulated as shown in (\ref{eqn:kl_approx}) between the words $w$ and $v$.
    \item \textbf{KL$\_$comp}: Maximum component-wise negative KL between words $w$ and $v$:
    \begin{eqnarray}&\emph{KL\_comp(w,v)} = \nonumber \\  \nonumber  & \max_{i,j=1,2,...,C} -KL(f_{w,i} (\bx)||f_{v,j} (\bx))  \end{eqnarray}
\end{enumerate}

Table~\ref{table:scws} compares the performance of the approaches on the SCWS dataset. It is evident from Table~\ref{table:scws} that GM$\_$KL achieves better correlation than existing approaches for various metrics on SCWS dataset.

\begin{table}[ht]
\caption{\textbf{Spearman correlation ({$\rho$ * 100}) on SCWS.}}
\begin{tabular}{|l|l|l|l|}
\hline
\textbf{Metric} & \textbf{w2g} & \textbf{w2gm} & \textbf{GM$\_$KL}  \\ \hline
 MaxCos   &     45.48       &    54.95  &   \textbf{55.09}    \\ \hline
 AvgCos   &     45.48       &    54.78  &   \textbf{57.48}    \\ \hline
KL\_approx    &    39.16      &   37.42  &   \textbf{48.06}     \\ \hline
KL\_comp    &    26.81      &  30.20 & \textbf{35.62} \\ \hline
\end{tabular}
\label{table:scws}
\end{table}

Table \ref{table:corr_res} shows the Spearman correlation values of GM$\_$KL model evaluated on the benchmark word similarity datasets: SL \citep{hill2015simlex}, WS, WS-R, WS-S \citep{finkelstein2002placing}, MEN
\citep{bruni2014multimodal}, MC
\citep{miller1991contextual}, RG \citep{rubenstein1965contextual}, YP \citep{yang2006verb}, MTurk-287 and MTurk-771 \citep{radinsky2011word, halawi2012large}, and RW \citep{luong2013better}. The metric used for comparison is 'AvgCos'. 
It can be seen  that for most of the datasets, GM$\_$KL  achieves significantly better correlation score than w2g and w2gm approaches. Other datasets such as  MC and RW  consist of only a single sense, and hence w2g model performs better and GM$\_$KL achieves next better performance. The YP dataset have multiple senses but does not contain entailed data and hence could not make use of entailment benefits of GM$\_$KL.\\
\begin{table}[ht]
\caption{\textbf{Spearman correlation results on word similarity datasets.}}
\begin{tabular}{|p{2cm}|p{1.5cm}|p{1.5cm}|p{1.5cm}|}
\hline
\textbf{\begin{tabular}[c]{@{}l@{}}Dataset\end{tabular}} & \textbf{\begin{tabular}[c]{@{}l@{}}w2g\end{tabular}} & \textbf{\begin{tabular}[c]{@{}l@{}}w2gm\end{tabular}} & \textbf{\begin{tabular}[c]{@{}l@{}}GM$\_$KL\\(Ours)\end{tabular}} \\ \hline
SL                                               & 14.29                                                      & 19.77                                                       & \textbf{22.96}                                              
\\ \hline
WS                                                          & 47.63                                                      & 58.35                                                       & \textbf{64.79}                 
\\ \hline
\begin{tabular}[c]{@{}l@{}}WS-S\end{tabular}             & 49.43                                                      & 59.22                                                       & \textbf{65.48}                 
\\ \hline
\begin{tabular}[c]{@{}l@{}}WS-R\end{tabular}             & 47.85                                                      & 56.90                                                       & \textbf{64.67}                     
\\ \hline
MEN                                                         & 42.61                                                      & 55.96                                                       & \textbf{57.04}                 
\\ \hline
MC                                                          & 43.01                                       & 39.21                                                       & 41.05                                         
\\ \hline
RG                                                          & 27.13                                                      & 49.68                                                       & \textbf{51.87}                 
\\ \hline
YP                                                          & 12.05                                                      & 28.74                                                 & 21.50                                
\\ \hline
\begin{tabular}[c]{@{}l@{}}MT-287\end{tabular}           & 51.41                                                      & 61.25                                                       & \textbf{64.00}                 
\\ \hline
\begin{tabular}[c]{@{}l@{}}MT-771\end{tabular}           & 41.38                                                      & 50.58                                                       & \textbf{51.68}                 
\\ \hline
RW                                                          & 18.43                                       & 12.65                                                       & 12.96                                         
\\ \hline
\end{tabular}
\label{table:corr_res}
\end{table}

Table \ref{table:entail_res} shows the evaluation results of GM$\_$KL model on the entailment datasets such as entailment pairs dataset \citep{baroni2012entailment} created from WordNet with both positive and negative labels, a crowdsourced dataset \citep{turney2015experiments}  of $79$ semantic relations labelled as entailed or not and annotated distributionally similar nouns dataset \citep{kotlerman2010directional}. The 'MaxCos' similarity metric is used for evaluation and the best precision and best F1-score is shown, by picking the optimal threshold. Overall, GM$\_$KL performs  better than both w2g and w2gm approaches.
\begin{table}[ht]
\caption{\textbf{Results on entailment datasets}}
\begin{tabular}{|p{15mm}|p{15mm}|p{10mm}|p{10mm}|p{12mm}|}
\hline
\textbf{Dataset}                                 & \textbf{Metric} & \textbf{w2g} & \textbf{w2gm} & \textbf{\begin{tabular}[c]{@{}l@{}}GM$\_$KL\\(Ours)\end{tabular}} \\ \hline
\cite{turney2015experiments}    & Precision       & 51.69        & 53.47         & \textbf{54.25}  \\ \hline
                                                 & F1        & 65.41        & 66.27         & \textbf{66.32}  \\ \hline
\cite{baroni2012entailment}     & Precision     & 57.18        & 66.42         & \textbf{67.55}  \\ \hline
                                                 & F1        & 63.72        & 70.72         & \textbf{71.49}  \\ \hline
\cite{kotlerman2010directional} & Precision       & 66.12        & 69.89         & \textbf{70.00}  \\ \hline
                                                 & F1        & 46.07        & 46.40         & \textbf{47.48}           \\ \hline
\end{tabular}
\label{table:entail_res}
\end{table}

\section{Conclusion}
We proposed a  KL divergence based energy function for  learning multi-sense word embedding distributions modelled as Gaussian mixtures. Due to the intractability of the Gaussian mixtures for the KL divergence measure, we use an approximate KL divergence function. We also demonstrated that the  proposed GM$\_$KL approaches performed better than other approaches on the benchmark word similarity and entailment datasets.

\bibliography{word_embed}
\bibliographystyle{acl_natbib}

\newpage
\appendix
\addcontentsline{toc}{section}{Appendices}
\section*{Supplementary Material}
\section{Approximation for KL divergence between mixtures of gaussians}

KL between gaussian mixtures $f_{w}(\bx)$ and $f_{v}(\bx)$  can be decomposed as: 
\begin{equation}
    \begin{aligned}
    KL(f_{w}(\bx)||f_{v}(\bx)) &= LB_{f_{w}}(f_{w}(\bx)) - LB_{f_{w}}(f_{v}(\bx)) \\
    LB_{f_{w}}(f_{v}(\bx)) &= E_{x \sim f_{w}}[log f_{v}(\bx)] \\
    LB_{f_{w}}(f_{w}(\bx)) &= E_{x \sim f_{w}}[log f_{w}(\bx)]
    \end{aligned}
\end{equation}

\citep{hershey2007approximating} presents KL approximation between gaussian mixtures using
\begin{enumerate}
\item product of gaussian approximation method where KL is approximated using product of component gaussians and 
\item variational approximation method where KL is approximated by introducing some variational parameters.\\
\end{enumerate}

The product of component gaussian approximation method using Jensen's inequality provides upper bounds as shown in equations \ref{eqn:pdt_of_gauss} and \ref{eqn:pdt_of_gaussi}.
\begin{eqnarray}
\label{eqn:pdt_of_gauss}
    LB_{f_{w}}(f_{v}(\bx)) \leq \sum_{i} p_{w,i}~log (\sum_{j} p_{v,j}~EL_{ij}(w,v)) \\
\label{eqn:pdt_of_gaussi}
    LB_{f_{w}}(f_{w}(\bx)) \leq \sum_{i} p_{w,i}~log (\sum_{k} p_{w,k}~EL_{ik}(w,w))
\end{eqnarray}

The variational approximation method provides lower bounds as shown in equations \ref{eqn:var_approx} and \ref{eqn:var_approxi}.
\begin{equation}
\label{eqn:var_approx}
\begin{aligned}
LB_{f_{w}}(f_{v}(\bx)) \geq \\ {\sum_i} p_{w,i} \log{{\sum_j p_{v,j} \exp({-KL(f_{w,i}(\bx)||f_{v,j}(\bx))})}} -\\ \sum_i p_{w,i} H(f_{w,i}(\bx))
\end{aligned}
\end{equation}

\begin{equation}
\label{eqn:var_approxi}
\begin{aligned}
LB_{f_{w}}(f_{w}(\bx)) \geq \\ {\sum_i} p_{w,i} \log{\sum_k
p_{w,k} \exp({-KL(f_{w,i} (\bx)||f_{w,k}(\bx))})} - \\ \sum_i p_{w,i} H(f_{w,i}(\bx))
\end{aligned}
\end{equation}

where $H$ represents the entropy term and the entropy of $i^{th}$ component of word $w$ with dimension $D$ is given as
\[
H(f_{w,i}(\bx))\triangleq -\int_x f_{w,i}(\bx)log{f_{w,i}(\bx)}dx = \frac{1}{2}log{((2\pi e)^D|\Sigma_{w,i}|)} 
\]

In \citep{durrieu2012lower}, the authors combine the lower and upper bounds from  approximation methods of \citep{hershey2007approximating} to formulate a stricter bound on KL between gaussian mixtures. \\

From equations \ref{eqn:pdt_of_gauss} and \ref{eqn:var_approxi}, a stricter lower bound for KL between gaussian mixtures is obtained as shown in equation \ref{eqn:lb}
\begin{equation}
\label{eqn:lb}
\begin{aligned}
KL_{lower}(f_{w}(\bx)||f_{v}(\bx)) &\geq \\ {\sum_i} p_{w,i} \log{\frac{\sum_k
p_{w,k} \exp({-KL(f_{w,i} (\bx)||f_{w,k}(\bx))})}{\sum_j p_{v,j} EL_{ij}(w,v)}} - \\ \sum_i p_{w,i} H(f_{w,i}(\bx))
\end{aligned}
\end{equation}

From equations \ref{eqn:pdt_of_gaussi} and \ref{eqn:var_approx}, a stricter upper bound for KL between gaussian mixtures is obtained as shown in equation \ref{eqn:ub}
\begin{equation}
\label{eqn:ub}
\begin{aligned}
KL_{upper}(f_{w}(\bx)||f_{v}(\bx)) &\leq \\ {\sum_i} p_{w,i} \log{\frac{\sum_k p_{w,k} EL_{ik}(w,w)}{\sum_j p_{v,j} \exp({-KL(f_{w,i}(\bx)||f_{v,j}(\bx))})}} +\\ \sum_i p_{w,i} H(f_{w,i}(\bx))
\end{aligned}
\end{equation}\\ \\
Finally, the KL between gaussian mixtures is taken as the mean of KL upper and lower bounds as shown in equation \ref{eqn:kl_approx_new}.
\begin{equation}
\label{eqn:kl_approx_new}
\begin{aligned}
KL(f_{w}(\bx)||f_v(\bx)) \triangleq [KL_{upper}(f_{w}(\bx)||f_{v}(\bx)) + \\ KL_{lower}(f_{w}(\bx)||f_{v}(\bx))]/2 
\end{aligned}
\end{equation}
\end{document}